%% file: main.tex
\documentclass{vgtc}                          

\graphicspath{{figures/}{pictures/}{images/}{./}} 

\usepackage{times}                     

\usepackage{tabu}                      
\usepackage{booktabs}                  
\usepackage{lipsum}                    
\usepackage{mwe}                       
\usepackage{comment}                   

\usepackage[numbers]{natbib} 

\usepackage{mathptmx}                  

\input{macro}

\onlineid{0}

\vgtccategory{Research}

\vgtcinsertpkg

\title{Designing Memory-Augmented AR Agents for Spatiotemporal Reasoning in Personalized Task Assistance}

\author{Dongwook Choi\thanks{e-mail: dwchoi0610@gmail.com}\\
    \scriptsize Language \& AGI Lab\\Yonsei University %
\and Taeyoon Kwon\thanks{e-mail: kwonconnor101@yonsei.ac.kr}\\
    \scriptsize Language \& AGI Lab\\Yonsei University %
\and Dongil Yang\thanks{e-mail: wingu@yonsei.ac.kr}\\
    \scriptsize Language \& AGI Lab\\Yonsei University %
\and Hyojun Kim\thanks{e-mail: magichjkim12@gmail.com}\\
    \scriptsize Language \& AGI Lab\\Yonsei University %
\and Jinyoung Yeo\thanks{e-mail: jinyeo@yonsei.ac.kr (correspondence)}\\
    \scriptsize Language \& AGI Lab\\Yonsei University
}

\input{latex/0_abstract}


\begin{document}


\input{latex/1_intro}
\input{latex/2_related}
\input{latex/3_scenario_setup}
\input{latex/4_system}
\input{latex/5_impl}
\input{latex/6_use}

\input{latex/7_conclusion}

\acknowledgments{
This work was supported by STEAM R\&D Project, NRF, Korea (RS-2024-00454458).}
\bibliographystyle{plainnat} 

\bibliography{main}
\end{document}

%% file: macro.tex
\newcommand{\eg}{{\it e.g.}}

%% file: latex/0_abstract.tex
\abstract{
Augmented Reality (AR) systems are increasingly integrating foundation models, such as Multimodal Large Language Models (MLLMs), to provide more context-aware and adaptive user experiences. 
This integration has led to the development of AR agents to support intelligent, goal-directed interactions in real-world environments.
While current AR agents effectively support immediate tasks, they struggle with complex multi-step scenarios that require understanding and leveraging user's long-term experiences and preferences.
This limitation stems from their inability to capture, retain, and reason over historical user interactions in spatiotemporal contexts. 
To address these challenges, we propose a conceptual framework for memory-augmented AR agents that can provide personalized task assistance by learning from and adapting to user-specific experiences over time. 
Our framework consists of four interconnected modules: (1) Perception Module for multimodal sensor processing, (2) Memory Module for persistent spatiotemporal experience storage, (3) Spatiotemporal Reasoning Module for synthesizing past and present contexts, and (4) Actuator Module for effective AR communication. 
We further present an implementation roadmap, a future evaluation strategy, a potential target application and use cases to demonstrate the practical applicability of our framework across diverse domains.
We aim for this work to motivate future research toward developing more intelligent AR systems that can effectively bridge user's interaction history with adaptive, context-aware task assistance.
}

%% file: latex/1_intro.tex
\firstsection{Introduction}

\maketitle

Augmented Reality (AR) is an innovative technology that enhances the real world with virtual content aligned in 3D space, enabling users to perceive and interact with both physical and digital elements simultaneously~\cite{azuma1997survey, milgram1994taxonomy}.
Conventional AR systems primarily focus on leveraging contextual cues\textemdash such as user's gaze, mobility, and interaction context\textemdash to provide relevant and adaptive augmented content to the real world through Head-Mounted Displays (HMDs)~\cite{davari2024towards, grubert2016towards}.
Building on this foundation, there is a growing interest in integrating Generative AI into AR systems to support more context-aware and adaptive user experiences~\cite{ castelo2023argus, dogan2024augmented, shi2025caring}.

Such AI integration has become feasible due to the advancement of foundation models\textemdash such as Large Language Models (LLMs) and Multimodal LLMs (MLLMs)\textemdash which show exceptional capabilities in commonsense reasoning, multimodal understanding, and adaptive decision-making capabilities. 
Thereby, these foundation models are being increasingly utilized as autonomous agents capable of decision-making and environmental interaction, as demonstrated in applications ranging from web navigation agents~\cite{chae2024web, chae2025web, he2024webvoyager} to embodied robotics agents~\cite{ahn2022can, driess2023palm, kwon2025embodied, zitkovich2023rt}. 
Following this approach, in the AR domain, recent research has explored AR agents that leverage foundation models to support more intelligent and goal-directed interaction~\cite{castelo2023argus, dogan2024augmented}. 
These AR agents excel at grounding real-time visual contexts with language instructions, allowing them to interpret and act upon complex scenarios~\cite{ pei2024autonomous, wu2024artist, xu2024augmented}.

However, as shown in Figure~\ref{fig:sample_1}, while current AR agents are effective in supporting immediate tasks, they fall short in capturing and reusing users' long-term experiences~\cite{haddad2025ar,li2025satori,yang2025socialmind}. 
As a result, they struggle to assist with complex multi-step task contexts grounded in personal experience (\eg, reproducing a user's cooking routine with ingredient-specific preferences or organizing items based on prior user-defined storage configurations), which limits their ability to provide truly personalized and contextually relevant assistance that builds upon users' historical interactions and preferences. 
Therefore, we argue that \textbf{memory-augmented AR agents} are essential for providing personalized assistance by recalling, reasoning over, and adapting to user-specific experiences.
Yet, designing memory-augmented AR agents introduces several key challenges: multimodal perception under uncertainty, persistent memory management, spatiotemporal reasoning, and effective action presentation in AR environments.

\input{figures_tex/Figure_1}

To address these challenges, we propose a conceptual framework for memory-augmented AR agents that support personalized task assistance, organized around four interconnected modules. 
(1) Perception Module: comprehensive processing and integration of multimodal sensor information to generate structured representations of the user's current context. 
(2) Memory Module: persistent preservation of spatiotemporal user experiences in procedural formats, enabling contextual retrieval beyond simple linguistic matching to incorporate spatial configurations and behavioral patterns. 
(3) Spatiotemporal Reasoning Module: synthesis of past experiences with current observations to recognize procedural states, track multi-step task progress, and infer next-step guidance while resolving noisy or partial perception through contextual alignment. 
(4) Actuator Module: effective communication and execution of agent decisions within the AR environment. 
This framework emphasizes the functional roles and interactions of these modules, offering a foundation for building agents that adapt to user-specific routines, spatial configurations, and task sequences.

To summarize, this paper presents a conceptual framework for memory-augmented AR agents, proposing a modular approach to address current limitations in personalized task assistance. 
Our contributions include: (1) identifying key design challenges of memory-augmented AR agents, (2) establishing a comprehensive framework supported by an implementation roadmap and evaluation strategy, and (3) demonstrating potential use cases for future applications.

%% file: figures_tex/Figure_1.tex
\begin{figure}[t]
  \centering
  \includegraphics[width=0.8\linewidth]{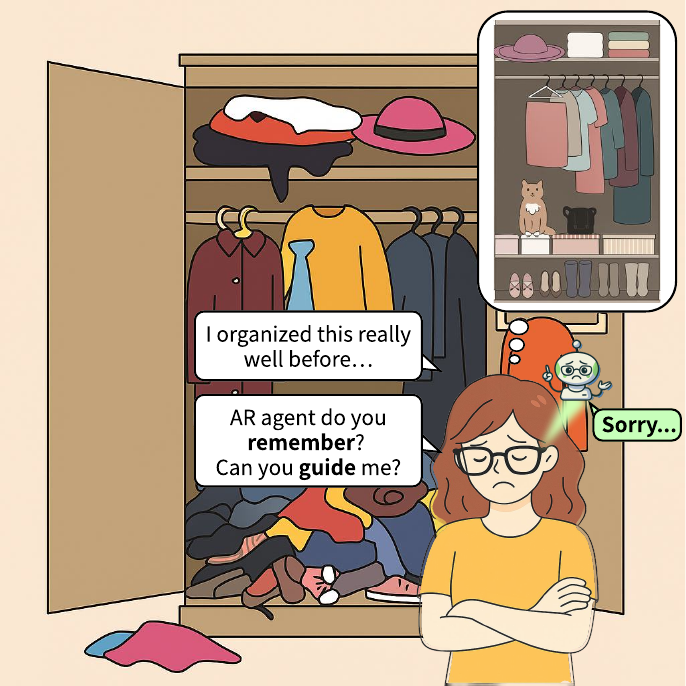}
  \caption{A motivating example of a user asking an AR agent for guidance based on a prior organization experience. 
  The agent fails to leverage user-specific memory, revealing a key limitation of current AR systems and underscoring the need for memory-augmented AR agents that support personalized task assistance.}
  \label{fig:sample_1}
\end{figure}

%% file: latex/2_related.tex
\section{Related Works}

\subsection{Context-aware AR Assistant}
AR systems have long aimed to assist users by interpreting and responding to contextual information in real-world environments. 
Early context-aware AR systems relied on predefined markers~\cite{kato1999marker, liu2021virtual, zauner2003authoring}, geolocation~\cite{shea2017location, takacs2008outdoors}, or simple object detection~\cite{hu2020object, kastner2021integrative, malta2021augmented} to trigger contextual responses.
While these systems could recognize specific objects or locations and overlay relevant information, they lacked a sophisticated understanding of dynamic environmental conditions or user states.
With the advent of large-scale foundation models, modern context-aware AR assistants increasingly integrate LLMs and MLLMs to enable more intelligent and adaptive support. These systems can provide knowledge grounded in observed objects and scenes~\cite{dogan2024augmented}, give proactive assists~\cite{li2025satori, yang2025socialmind}, and generate procedural guidance tailored to task progress~\cite{castelo2023argus, pei2024autonomous, wu2024artist, xu2024augmented}. 
Collectively, these studies illustrate how AR assistants have evolved to incorporate multimodal understanding and proactive assistance.

\subsection{Multimodal Scene Graph Generation}
Scene graphs have emerged as a powerful abstraction for representing complex environments by unifying multimodal perceptual input. Recent advances in Multimodal Scene Graph Generation (MSGG) extend this abstraction beyond static 2D scenes by incorporating temporal cues, 3D geometric reasoning, and language understanding~\cite{cong2021spatial, feng20233d, zhou2024openpsg}. These works enable richer representations that are particularly well suited for interactive and dynamic settings like AR.
In the 3D domain, scene graphs built from point cloud data allow for more precise spatial grounding, while transformer-based architectures enhance the modeling of complex relational structures across modalities~\cite{lv2024sgformer, sarkar2025crossover}. Open-vocabulary frameworks further support adaptation to real-world environments without predefined label constraints~\cite{koch2024open3dsg}. Additionally, recent research has explored scene graphs as interfaces for reasoning within LLMs, demonstrating their potential as shared representations for downstream tasks~\cite{yang2025llm}.
These works highlight the effectiveness of scene graphs as a unified, interpretable, and extensible structure for multimodal integration and reasoning, motivating their use in adaptive AR systems.

\subsection{Memory-Augmented Agent}
Recent advances in memory-augmented agents have enabled more adaptive and context-aware interactions across various domains ~\cite{sumers2023cognitive}.
In Embodied agents, research primarily focuses on maintaining structured episodic memories and semantic contexts to support complex and interactive tasks~\cite{kwon2025embodied, wang2024karma, zhu2024retrieval}. Especially,~\citet{kwon2025embodied} highlights the importance of distinguishing and utilizing user-specific semantic knowledge and routine patterns for effective personalized assistance.
Web and GUI-based agents leverage memory specifically designed to retain personalized user data, interaction history, and individual user preferences~\cite{cai2025large, wang2024agent}.
Recent AR research has investigated memory-augmented systems for delivering contextual assistance in real-time environments~\cite{haddad2025ar, li2025satori}.
Typical systems provide personalized assistance by recognizing immediate user contexts and recalling recent interactions through AR glasses.
However, their focus remains on short-term interactions, offering limited support when users seek to reproduce complex, personalized routines grounded in long-term experiences.

%% file: latex/3_scenario_setup.tex
\section{Scenario Setup and Memory Construction}
We consider a two-phase interaction scenario between the user and the AR system. 
In the first phase, the user records their everyday activities (\eg, cooking a recipe or organizing a space) using an AR glass, then the user names each recording with a personalized title, such as “\textit{Mom's Chicken Stew Recipe}”. 
These titled recordings are later transformed into structured, task-relevant memory representations via offline processing of egocentric video and associated sensor data (\eg, hand pose, audio)~\cite{ji2020action, liu2020beyond, teng2021target}.

In the second phase, when the user returns to the same location (\eg, say, standing once again in front of their kitchen counter) and verbally specifies which episode to recall, which initiates the retrieval process.
The titles of retrieved previously recorded episodes can appear as a subtle overlay in the user's view, offering them a chance to recall their past approach.
Once the user selects the memory, AR system uses the recorded episode as a reference to assess the user's current action and environment, tracking their progress and suggesting the next step.
Instead of suggesting a common next step, the system provides assistance based on the user's past experience, using visual or audio cues.
For example, when preparing ingredients for a meal\textemdash as in the previously saved episode titled “\textit{Mom's Chicken Stew Recipe}”\textemdash the AR display may indicate precisely which item to use next or how it was previously handled, based on how the user prepared the dish before. 
In this way, the system helps the user recall and follow their own personalized workflow with context-aware assistance.

%% file: latex/4_system.tex
\input{figures_tex/Figure_2}

\section{Memory-augmented AR Agent Framework}
To enable personalized task assistance in real-world environments, we propose a memory-augmented AR assistant framework that leverages user's past experiences in physical environments.

\subsection{Framework Overview}
As illustrated in Figure~\ref{fig:sample_2}, our framework operates over a two-phase interaction.
In the Recording phase, users record everyday activities using AR glasses; these recordings are later transformed offline into structured, task-relevant memory representations and stored as episodic memory.
In the Recall phase, the following four interconnected modules work together to provide personalized assistance grounded in episodic memory: \textbf{(1) Perception Module}, for understanding and structuring the current user's context by processing multimodal sensory inputs; \textbf{(2) Memory Module}, for maintaining and organizing previously stored episodic memories as retrievable references; \textbf{(3) Spatiotemporal Reasoning Module}, for aligning current context with past experiences to infer user goals, estimate task progress, and suggest context-aware next steps to support assistance; and \textbf{(4) Actuator Module}, for deciding and providing user-facing assistance grounded in the plan and current context.
Together, these components form a closed-loop system that continuously adapts to the user's behavior and task environment.

\subsection{Unified Representation: Scene Graph}
\subsubsection{Motivation}
To support adaptive assistance in complex AR environments, we adopt the \textbf{scene graph} as a unified, structured representation of the user's surroundings.
Scene graphs provide a flexible abstraction that integrates multimodal perceptual information\textemdash such as objects, spatial layouts, actions, and interactions\textemdash into a coherent structure.
Also, serving as a shared data representation across all of the modules, scene graphs promote consistent communication and reduce the overhead of modality-specific integration.
Moreover, recent work~\cite{yang2025llm} demonstrates that LLMs are capable of understanding and reasoning over scene graphs, supporting our design choice.

\subsubsection{Scene Graph Representation}
We represent the user's surroundings at timestep $t$ using a dynamic scene graph $G_t = (V_t, E_t)$, where nodes $v_i \in V_t$ denote entities such as objects, user hands, actions, and UI elements. 
Each node is associated with multimodal features $f_i$ derived from egocentric visual input, gaze, hand pose, and speech, capturing both its type and its context in the current scene.
Directed edges $e_{ij} \in E_t$ may represent observed physical interactions (\eg, grasping, next to), inferred attentional cues (\eg, attending to, looking at), or planned guidance relations (\eg, find, notify, to be grasped).

\subsection{Perception Module}

The perception module constructs a structured representation of the user's current surroundings based on multimodal sensory inputs. 
At each timestep, the system captures egocentric data from the AR glasses, including visual scenes, hand pose, gaze direction, and audio signals such as speech. 
The module integrates them to generate a unified scene graph representation of the current context.

This is achieved by leveraging recent advances in MLLMs, which have demonstrated the capability to generate scene graph structures directly from complex, multimodal inputs.~\cite{ji2020action, liu2020beyond, teng2021target}
By employing these models, the perception module translates raw sensory observations into structured entities, spatial relations, and interaction cues\textemdash forming the basis for downstream modules.

\subsection{Memory Module}
The memory module stores structured representations of the user's past task experiences. 
Each memory consists of a sequence of scene graphs constructed from previously recorded experiences, capturing key object interactions, spatial layouts, and procedural steps.
These structured experiences are later referenced by the Spatiotemporal Reasoning module, which support goal-directed assistance, enabling the system to guide the user through tasks based on their own prior workflows.

\subsection{Spatiotemporal Reasoning Module}
We define the Spatiotemporal Reasoning module as consisting of three core components: (1) \textit{Task Intent Inference}, (2) \textit{Task Progress Tracking}, and (3) \textit{Action Planning}. 

\subsubsection{Task Intent Inference}

This component analyzes a stored memory episode\textemdash represented as a temporal sequence of scene graphs\textemdash to infer the user's original task intent and their personalized procedure for completing it. 
By examining the temporal sequence of user interactions, it identifies key sub-goals, intermediate steps, and habitual action patterns that define the user's approach.
This enables the agent to align current behavior with personalized task representations and support user-specific task execution.

\subsubsection{Task Progress Tracking}
This component tracks the user's progress within an ongoing task by analyzing sequential egocentric observations represented as scene graphs, which are continuously streamed from the perception module. 
Rather than recognizing isolated actions, it interprets short-term behavioral patterns\textemdash maintained in working memory\textemdash and aligns them with expected procedural steps drawn from previously stored memories. 
By comparing current activity to personalized task flows, the module identifies the user's current stage and ensures continuous guidance. 
It distinguishes meaningful task-related actions from short-term off-task behaviors\textemdash such as answering a phone call or briefly stepping away\textemdash so that the system avoids false deviations and maintains reliable progress tracking even in noisy, real-world situations.

\subsubsection{Action Planning}
This component determines plausible next steps by reasoning over the current user context\textemdash represented as a scene graph from the \textit{Task Progress Tracking} component\textemdash and the user's personalized task plan\textemdash retrieved and interpreted from stored task memories via the \textit{Task Intent Inference} component.
It operates by constructing or modifying a scene graph that encodes the intended guidance for the user, including required actions or interventions.
This updated representation is then passed to the Actuator module.

\subsection{Actuator Module}
The Actuator module determines the final action to assist the user and presents it through appropriate modalities.
Based on the output scene graph from the Spatiotemporal Reasoning module, it selects the most relevant instruction or intervention to support task completion.
The module ensures that assistance is grounded in the current execution context by filtering out actions that are infeasible or irrelevant, considering commonsense constraints and the availability of supporting tools (\eg, object-level highlighting, brief on-screen tips, or voice-based cues)~\cite{zhang2024sgedit, zheng2024editroom}.

Recent work has demonstrated that LLMs can not only interpret multimodal context but also operate external tools and interfaces to provide assistance in situated environments~\cite{patil2024gorilla, qin2023toolllm}.
Inspired by these capabilities, our Actuator similarly interprets the scene graph to decide both what assistance to provide and how to execute it\textemdash choosing the most appropriate modality based on the user's current execution context.

%% file: figures_tex/Figure_2.tex
\begin{figure*}[t!]
  \centering
  \includegraphics[width=1\linewidth]{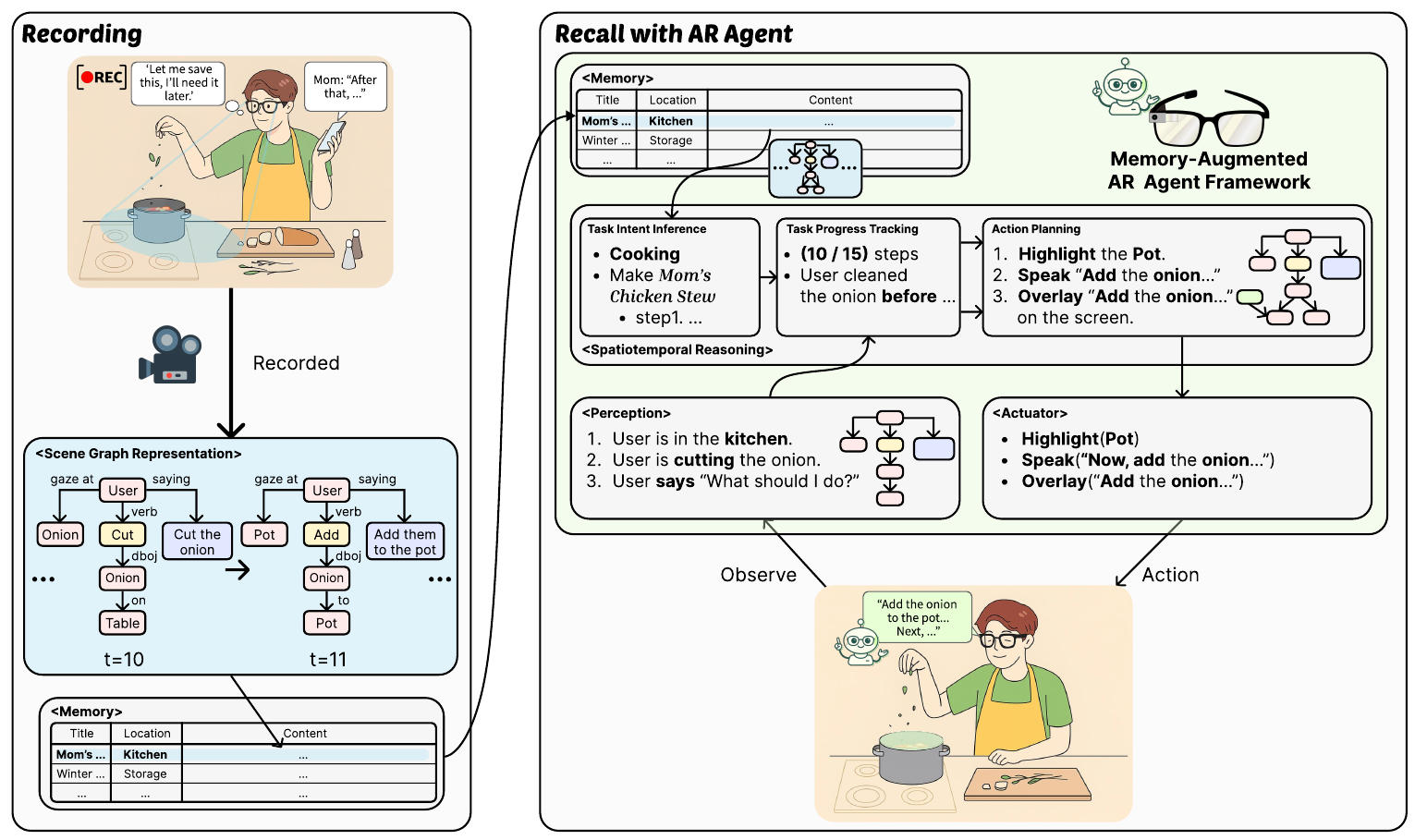}
  \caption{A two-phase conceptual framework for personalized memory-augmented AR agents. User experiences are recorded and encoded as scene graphs during the Recording phase. In the Recall phase, AR agents leverage these graphs to provide personalized guidance (\eg, highlighting the pot and saying “Add the onion…”) as the user prepares a personal recipe such as \textit{Mom’s Chicken Stew}.}
  \label{fig:sample_2}
\end{figure*}

%% file: latex/5_impl.tex
\section{Implementation Roadmap and Evaluation Strategy}

To make the proposed memory-augmented AR agent framework, as shown in Figure~\ref{fig:sample_2}, more implementable and practically grounded, we outline a high-level roadmap to implement and evaluate a memory-augmented AR agent.

\subsection{System Components for Implementation}

Each module of the proposed framework is instantiated using AR development tools and existing foundation models. 
In particular, we set the base simulation engine with Unity, enabling the generation of an effective AR environment, 
with GPT-4o-realtime~\cite{hurst2024gpt} serving as the primary reasoning and planning backbone across the system.
The Perception Module employs SAM2~\cite{ravi2024sam} for object and region detection, and processes multimodal inputs (\eg, gaze trajectories, hand poses, speech transcripts) through the primary backbone.
Considering deployment environments, this module may alternatively utilize open-source MLLMs fine-tuned on the Ego4D-EASG dataset~\cite{rodin2024action}.

During memory construction, recorded episodes are converted into temporal sequences of scene graphs using the same scene graph construction pipeline as in the Perception Module.
These are embedded with text-embedding models~\cite{neelakantan2022text} and stored in a vector database (\eg, FAISS~\cite{douze2024faiss, wang2024vector}) together with structured metadata including episode titles, timestamps, and locations. 
At the start of the recall phase, the most relevant episode is retrieved and loaded into working memory. 
The Spatiotemporal Reasoning Module aligns the current scene graph with recalled memory entities, tracks task progress, and generates action plans.

The Actuator Module receives an action plan in the form of a scene graph and parses it to determine the required interactions.
The backbone model is employed to accurately identify target objects or regions in the scene~\cite{zhang2024sgedit, zheng2024editroom}.
It then calls predefined actuation functions in Unity to execute the corresponding actions within the AR environment.

\subsection{Evaluation Plan}

A user study is conducted to evaluate the system's ability to provide effective and context-aware guidance in real-world tasks.
Evaluation metrics include: (1) Task Completion Rate, defined as the proportion of steps successfully completed with the provided guidance; 
(2) Task Completion Time, the total time required to complete each task; (3) NASA Task Load Index (NASA-TLX)~\cite{hart1988development}, measuring user workload across cognitive and physical dimensions; and (4) a User Satisfaction Survey assessing perceived usefulness, naturalness, and reliability of the system.
For comparison, participants are divided into two distinct user groups: (a) memory-augmented AR guidance, where the system provides personalized recall and context-aware instructions, and (b) Text-only AR guidance, where the AR glasses display fixed recipe instructions as on-screen text.
This design enables capturing both the objective task performance benefits and the subjective user experience improvements afforded by the proposed memory-augmented AR agent.

%% file: latex/6_use.tex
\section{Potential Target Application and Use Cases}
We present an initial target application followed by several potential use cases, demonstrating how our memory-augmented AR agent can be applied across diverse personalized, context-aware tasks. 

\subsection{Target Application: Memory-Assisted Cooking Recall}
As an initial prototype, we have chosen the domain of personalized cooking assistance, a task that naturally benefits from combining past user experiences with real-time perception and guidance~\cite{majil2022augmented,vir2024archef}.
Users often modify recipes and cooking flows based on personal tastes and habits\textemdash such as skipping marination or prepping all ingredients before heating the stove. 
Our system can recall these personalized workflows and provide context-aware prompts, helping users follow their own preferred cooking style consistently.

\subsection{Potential Use Cases}

We identify three representative scenarios where memory-augmented AR agents can provide tangible benefits:

\begin{itemize}
    \item \textbf{Routinely Organizing Household Items} \textemdash  From seasonal clothing storage to kitchen layout, users develop implicit organizational logic that's easily forgotten over time. 
    The agent retrieves previous configurations and helps restore or adapt personalized systems with minimal friction.
    
    \item \textbf{Repeating Personalized Health Training} \textemdash  In physical rehab or yoga, users find routines that best suit their bodies.
    By recording motion, setup, and pacing, the agent enables faithful repetition of these effective sessions\textemdash especially when reinitiating after breaks.
    This direction aligns with recent efforts~\cite{jo2023flowar}, which explore AR-based support for personalized home fitness.
    
    \item \textbf{Repeating a Personalized Experiment} \textemdash  Researchers often revisit past experiments with minor changes. 
    Remembering exact setups is difficult without detailed notes. 
    The agent references past layouts and sequences\textemdash like reagent order or labeling quirks\textemdash to support reproducibility and reduce error.
    Recent work has also explored AR-based support for laboratory training and experimental guidance~\cite{hallmann2023supporting,ismael2024acceptance}, reinforcing the relevance of such systems for improving safety and procedural accuracy.

\end{itemize}

%% file: latex/7_conclusion.tex
\section{Conclusion}
In this paper, we propose a conceptual framework for designing memory-augmented AR agents capable of providing personalized assistance through iterative perception and reasoning. 
By structuring user interactions into scene graph memories and aligning them with real-time context, our system supports adaptive guidance grounded in user-specific workflows. 
We hope this work stimulates further research into personalized, memory-augmented AR systems that bridge interaction history with context-aware task assistance.

%% file: main.bbl
\begin{thebibliography}{59}
\providecommand{\natexlab}[1]{#1}
\providecommand{\url}[1]{\texttt{#1}}
\expandafter\ifx\csname urlstyle\endcsname\relax
  \providecommand{\doi}[1]{doi: #1}\else
  \providecommand{\doi}{doi: \begingroup \urlstyle{rm}\Url}\fi

\bibitem[Ahn et~al.(2022)Ahn, Brohan, Brown, Chebotar, Cortes, David, Finn, Fu, Gopalakrishnan, Hausman, et~al.]{ahn2022can}
Michael Ahn, Anthony Brohan, Noah Brown, Yevgen Chebotar, Omar Cortes, Byron David, Chelsea Finn, Chuyuan Fu, Keerthana Gopalakrishnan, Karol Hausman, et~al.
\newblock Do as i can, not as i say: Grounding language in robotic affordances.
\newblock \emph{arXiv preprint arXiv:2204.01691}, 2022.

\bibitem[Azuma(1997)]{azuma1997survey}
Ronald~T Azuma.
\newblock A survey of augmented reality.
\newblock \emph{Presence: teleoperators \& virtual environments}, 6\penalty0 (4):\penalty0 355--385, 1997.

\bibitem[Cai et~al.(2025)Cai, Li, Wang, Zhu, Shen, Li, and Chua]{cai2025large}
Hongru Cai, Yongqi Li, Wenjie Wang, Fengbin Zhu, Xiaoyu Shen, Wenjie Li, and Tat-Seng Chua.
\newblock Large language models empowered personalized web agents.
\newblock In \emph{Proceedings of the ACM on Web Conference 2025}, pages 198--215, 2025.

\bibitem[Castelo et~al.(2023)Castelo, Rulff, McGowan, Steers, Wu, Chen, Roman, Lopez, Brewer, Zhao, et~al.]{castelo2023argus}
Sonia Castelo, Joao Rulff, Erin McGowan, Bea Steers, Guande Wu, Shaoyu Chen, Iran Roman, Roque Lopez, Ethan Brewer, Chen Zhao, et~al.
\newblock Argus: Visualization of ai-assisted task guidance in ar.
\newblock \emph{IEEE Transactions on Visualization and Computer Graphics}, 30\penalty0 (1):\penalty0 1313--1323, 2023.

\bibitem[Chae et~al.(2024)Chae, Kim, Ong, Gwak, Song, Kim, Kim, Lee, and Yeo]{chae2024web}
Hyungjoo Chae, Namyoung Kim, Kai Tzu-iunn Ong, Minju Gwak, Gwanwoo Song, Jihoon Kim, Sunghwan Kim, Dongha Lee, and Jinyoung Yeo.
\newblock Web agents with world models: Learning and leveraging environment dynamics in web navigation.
\newblock \emph{arXiv preprint arXiv:2410.13232}, 2024.

\bibitem[Chae et~al.(2025)Chae, Kim, Cho, Kim, Moon, Hwangbo, Lim, Kim, Hwang, Gwak, et~al.]{chae2025web}
Hyungjoo Chae, Sunghwan Kim, Junhee Cho, Seungone Kim, Seungjun Moon, Gyeom Hwangbo, Dongha Lim, Minjin Kim, Yeonjun Hwang, Minju Gwak, et~al.
\newblock Web-shepherd: Advancing prms for reinforcing web agents.
\newblock \emph{arXiv preprint arXiv:2505.15277}, 2025.

\bibitem[Cong et~al.(2021)Cong, Liao, Ackermann, Rosenhahn, and Yang]{cong2021spatial}
Yuren Cong, Wentong Liao, Hanno Ackermann, Bodo Rosenhahn, and Michael~Ying Yang.
\newblock Spatial-temporal transformer for dynamic scene graph generation.
\newblock In \emph{Proceedings of the IEEE/CVF international conference on computer vision}, pages 16372--16382, 2021.

\bibitem[Davari and Bowman(2024)]{davari2024towards}
Shakiba Davari and Doug~A Bowman.
\newblock Towards context-aware adaptation in extended reality: A design space for xr interfaces and an adaptive placement strategy.
\newblock \emph{arXiv preprint arXiv:2411.02607}, 2024.

\bibitem[Dogan et~al.(2024)Dogan, Gonzalez, Ahuja, Du, Cola{\c{c}}o, Lee, Gonzalez-Franco, and Kim]{dogan2024augmented}
Mustafa~Doga Dogan, Eric~J Gonzalez, Karan Ahuja, Ruofei Du, Andrea Cola{\c{c}}o, Johnny Lee, Mar Gonzalez-Franco, and David Kim.
\newblock Augmented object intelligence with xr-objects.
\newblock In \emph{Proceedings of the 37th Annual ACM Symposium on User Interface Software and Technology}, pages 1--15, 2024.

\bibitem[Douze et~al.(2024)Douze, Guzhva, Deng, Johnson, Szilvasy, Mazar{\'e}, Lomeli, Hosseini, and J{\'e}gou]{douze2024faiss}
Matthijs Douze, Alexandr Guzhva, Chengqi Deng, Jeff Johnson, Gergely Szilvasy, Pierre-Emmanuel Mazar{\'e}, Maria Lomeli, Lucas Hosseini, and Herv{\'e} J{\'e}gou.
\newblock The faiss library.
\newblock \emph{arXiv preprint arXiv:2401.08281}, 2024.

\bibitem[Driess et~al.(2023)Driess, Xia, Sajjadi, Lynch, Chowdhery, Wahid, Tompson, Vuong, Yu, Huang, et~al.]{driess2023palm}
Danny Driess, Fei Xia, Mehdi~SM Sajjadi, Corey Lynch, Aakanksha Chowdhery, Ayzaan Wahid, Jonathan Tompson, Quan Vuong, Tianhe Yu, Wenlong Huang, et~al.
\newblock Palm-e: An embodied multimodal language model.
\newblock 2023.

\bibitem[Feng et~al.(2023)Feng, Hou, Zhang, Wu, Guo, and Mian]{feng20233d}
Mingtao Feng, Haoran Hou, Liang Zhang, Zijie Wu, Yulan Guo, and Ajmal Mian.
\newblock 3d spatial multimodal knowledge accumulation for scene graph prediction in point cloud.
\newblock In \emph{Proceedings of the IEEE/CVF Conference on Computer Vision and Pattern Recognition}, pages 9182--9191, 2023.

\bibitem[Grubert et~al.(2016)Grubert, Langlotz, Zollmann, and Regenbrecht]{grubert2016towards}
Jens Grubert, Tobias Langlotz, Stefanie Zollmann, and Holger Regenbrecht.
\newblock Towards pervasive augmented reality: Context-awareness in augmented reality.
\newblock \emph{IEEE transactions on visualization and computer graphics}, 23\penalty0 (6):\penalty0 1706--1724, 2016.

\bibitem[Haddad et~al.(2025)Haddad, Wang, Shin, Liu, Wang, and Yu]{haddad2025ar}
Rapha{\"e}l A~El Haddad, Zeyu Wang, Yeonsu Shin, Ranyi Liu, Yuntao Wang, and Chun Yu.
\newblock Ar secretary agent: Real-time memory augmentation via llm-powered augmented reality glasses.
\newblock \emph{arXiv preprint arXiv:2505.11888}, 2025.

\bibitem[Hallmann et~al.(2023)Hallmann, Stechert, and Ahmed]{hallmann2023supporting}
Jona Hallmann, Carsten Stechert, and Syed Imad-Uddin Ahmed.
\newblock Supporting student laboratory experiments with augmented reality experience.
\newblock \emph{Proceedings of the Design Society}, 3:\penalty0 3235--3244, 2023.

\bibitem[Hart and Staveland(1988)]{hart1988development}
Sandra~G Hart and Lowell~E Staveland.
\newblock Development of nasa-tlx (task load index): Results of empirical and theoretical research.
\newblock In \emph{Advances in psychology}, volume~52, pages 139--183. Elsevier, 1988.

\bibitem[He et~al.(2024)He, Yao, Ma, Yu, Dai, Zhang, Lan, and Yu]{he2024webvoyager}
Hongliang He, Wenlin Yao, Kaixin Ma, Wenhao Yu, Yong Dai, Hongming Zhang, Zhenzhong Lan, and Dong Yu.
\newblock Webvoyager: Building an end-to-end web agent with large multimodal models.
\newblock \emph{arXiv preprint arXiv:2401.13919}, 2024.

\bibitem[Hu et~al.(2020)Hu, Weng, Chen, and Wang]{hu2020object}
Mingwei Hu, Dongdong Weng, Feng Chen, and Yongtian Wang.
\newblock Object detecting augmented reality system.
\newblock In \emph{2020 IEEE 20th International Conference on Communication Technology (ICCT)}, pages 1432--1438. IEEE, 2020.

\bibitem[Hurst et~al.(2024)Hurst, Lerer, Goucher, Perelman, Ramesh, Clark, Ostrow, Welihinda, Hayes, Radford, et~al.]{hurst2024gpt}
Aaron Hurst, Adam Lerer, Adam~P Goucher, Adam Perelman, Aditya Ramesh, Aidan Clark, AJ~Ostrow, Akila Welihinda, Alan Hayes, Alec Radford, et~al.
\newblock Gpt-4o system card.
\newblock \emph{arXiv preprint arXiv:2410.21276}, 2024.

\bibitem[Ismael et~al.(2024)Ismael, McCall, McGee, Belkacem, Stefas, Baixauli, and Arl]{ismael2024acceptance}
Muhannad Ismael, Roderick McCall, Fintan McGee, Ilyasse Belkacem, Micka{\"e}l Stefas, Joan Baixauli, and Didier Arl.
\newblock Acceptance of augmented reality for laboratory safety training: methodology and an evaluation study.
\newblock \emph{Frontiers in Virtual Reality}, 5:\penalty0 1322543, 2024.

\bibitem[Ji et~al.(2020)Ji, Krishna, Fei-Fei, and Niebles]{ji2020action}
Jingwei Ji, Ranjay Krishna, Li~Fei-Fei, and Juan~Carlos Niebles.
\newblock Action genome: Actions as compositions of spatio-temporal scene graphs.
\newblock In \emph{Proceedings of the IEEE/CVF conference on computer vision and pattern recognition}, pages 10236--10247, 2020.

\bibitem[Jo et~al.(2023)Jo, Seidel, Pahud, Sinclair, and Bianchi]{jo2023flowar}
Hye-Young Jo, Laurenz Seidel, Michel Pahud, Mike Sinclair, and Andrea Bianchi.
\newblock Flowar: How different augmented reality visualizations of online fitness videos support flow for at-home yoga exercises.
\newblock In \emph{Proceedings of the 2023 CHI Conference on Human Factors in Computing Systems}, pages 1--17, 2023.

\bibitem[K{\"a}stner et~al.(2021)K{\"a}stner, Eversberg, Mursa, and Lambrecht]{kastner2021integrative}
Linh K{\"a}stner, Leon Eversberg, Marina Mursa, and Jens Lambrecht.
\newblock Integrative object and pose to task detection for an augmented-reality-based human assistance system using neural networks.
\newblock In \emph{2020 IEEE Eighth International Conference on Communications and Electronics (ICCE)}, pages 332--337. IEEE, 2021.

\bibitem[Kato and Billinghurst(1999)]{kato1999marker}
Hirokazu Kato and Mark Billinghurst.
\newblock Marker tracking and hmd calibration for a video-based augmented reality conferencing system.
\newblock In \emph{Proceedings 2nd IEEE and ACM International Workshop on Augmented Reality (IWAR'99)}, pages 85--94. IEEE, 1999.

\bibitem[Koch et~al.(2024)Koch, Vaskevicius, Colosi, Hermosilla, and Ropinski]{koch2024open3dsg}
Sebastian Koch, Narunas Vaskevicius, Mirco Colosi, Pedro Hermosilla, and Timo Ropinski.
\newblock Open3dsg: Open-vocabulary 3d scene graphs from point clouds with queryable objects and open-set relationships.
\newblock In \emph{Proceedings of the IEEE/CVF Conference on Computer Vision and Pattern Recognition}, pages 14183--14193, 2024.

\bibitem[Kwon et~al.(2025)Kwon, Choi, Kim, Kim, Moon, Kwak, Huang, and Yeo]{kwon2025embodied}
Taeyoon Kwon, Dongwook Choi, Sunghwan Kim, Hyojun Kim, Seungjun Moon, Beong-woo Kwak, Kuan-Hao Huang, and Jinyoung Yeo.
\newblock Embodied agents meet personalization: Exploring memory utilization for personalized assistance.
\newblock \emph{arXiv preprint arXiv:2505.16348}, 2025.

\bibitem[Li et~al.(2025)Li, Wu, Chan, Turakhia, Castelo~Quispe, Li, Welch, Silva, and Qian]{li2025satori}
Chenyi Li, Guande Wu, Gromit Yeuk-Yin Chan, Dishita~Gdi Turakhia, Sonia Castelo~Quispe, Dong Li, Leslie Welch, Claudio Silva, and Jing Qian.
\newblock Satori: Towards proactive ar assistant with belief-desire-intention user modeling.
\newblock In \emph{Proceedings of the 2025 CHI Conference on Human Factors in Computing Systems}, pages 1--24, 2025.

\bibitem[Liu and Tanaka(2021)]{liu2021virtual}
Boyang Liu and Jiro Tanaka.
\newblock Virtual marker technique to enhance user interactions in a marker-based ar system.
\newblock \emph{Applied Sciences}, 11\penalty0 (10):\penalty0 4379, 2021.

\bibitem[Liu et~al.(2020)Liu, Jin, Xu, Gong, and Mu]{liu2020beyond}
Chenchen Liu, Yang Jin, Kehan Xu, Guoqiang Gong, and Yadong Mu.
\newblock Beyond short-term snippet: Video relation detection with spatio-temporal global context.
\newblock In \emph{Proceedings of the IEEE/CVF conference on computer vision and pattern recognition}, pages 10840--10849, 2020.

\bibitem[Lv et~al.(2024)Lv, Qi, Li, Yang, and Ma]{lv2024sgformer}
Changsheng Lv, Mengshi Qi, Xia Li, Zhengyuan Yang, and Huadong Ma.
\newblock Sgformer: Semantic graph transformer for point cloud-based 3d scene graph generation.
\newblock In \emph{Proceedings of the AAAI Conference on Artificial Intelligence}, volume~38, pages 4035--4043, 2024.

\bibitem[Majil et~al.(2022)Majil, Yang, and Yang]{majil2022augmented}
Isaias Majil, Mau-Tsuen Yang, and Sophia Yang.
\newblock Augmented reality based interactive cooking guide.
\newblock \emph{Sensors}, 22\penalty0 (21):\penalty0 8290, 2022.

\bibitem[Malta et~al.(2021)Malta, Mendes, and Farinha]{malta2021augmented}
Ana Malta, Mateus Mendes, and Torres Farinha.
\newblock Augmented reality maintenance assistant using yolov5.
\newblock \emph{Applied Sciences}, 11\penalty0 (11):\penalty0 4758, 2021.

\bibitem[Milgram and Kishino(1994)]{milgram1994taxonomy}
Paul Milgram and Fumio Kishino.
\newblock A taxonomy of mixed reality visual displays.
\newblock \emph{IEICE TRANSACTIONS on Information and Systems}, 77\penalty0 (12):\penalty0 1321--1329, 1994.

\bibitem[Neelakantan et~al.(2022)Neelakantan, Xu, Puri, Radford, Han, Tworek, Yuan, Tezak, Kim, Hallacy, et~al.]{neelakantan2022text}
Arvind Neelakantan, Tao Xu, Raul Puri, Alec Radford, Jesse~Michael Han, Jerry Tworek, Qiming Yuan, Nikolas Tezak, Jong~Wook Kim, Chris Hallacy, et~al.
\newblock Text and code embeddings by contrastive pre-training.
\newblock \emph{arXiv preprint arXiv:2201.10005}, 2022.

\bibitem[Patil et~al.(2024)Patil, Zhang, Wang, and Gonzalez]{patil2024gorilla}
Shishir~G Patil, Tianjun Zhang, Xin Wang, and Joseph~E Gonzalez.
\newblock Gorilla: Large language model connected with massive apis.
\newblock \emph{Advances in Neural Information Processing Systems}, 37:\penalty0 126544--126565, 2024.

\bibitem[Pei et~al.(2024)Pei, Viola, Huang, Wang, Ahsan, Ye, Yiming, Sai, Wang, Chen, et~al.]{pei2024autonomous}
Jiahuan Pei, Irene Viola, Haochen Huang, Junxiao Wang, Moonisa Ahsan, Fanghua Ye, Jiang Yiming, Yao Sai, Di~Wang, Zhumin Chen, et~al.
\newblock Autonomous workflow for multimodal fine-grained training assistants towards mixed reality.
\newblock \emph{arXiv preprint arXiv:2405.13034}, 2024.

\bibitem[Qin et~al.(2023)Qin, Liang, Ye, Zhu, Yan, Lu, Lin, Cong, Tang, Qian, et~al.]{qin2023toolllm}
Yujia Qin, Shihao Liang, Yining Ye, Kunlun Zhu, Lan Yan, Yaxi Lu, Yankai Lin, Xin Cong, Xiangru Tang, Bill Qian, et~al.
\newblock Toolllm: Facilitating large language models to master 16000+ real-world apis.
\newblock \emph{arXiv preprint arXiv:2307.16789}, 2023.

\bibitem[Ravi et~al.(2024)Ravi, Gabeur, Hu, Hu, Ryali, Ma, Khedr, R{\"a}dle, Rolland, Gustafson, et~al.]{ravi2024sam}
Nikhila Ravi, Valentin Gabeur, Yuan-Ting Hu, Ronghang Hu, Chaitanya Ryali, Tengyu Ma, Haitham Khedr, Roman R{\"a}dle, Chloe Rolland, Laura Gustafson, et~al.
\newblock Sam 2: Segment anything in images and videos.
\newblock \emph{arXiv preprint arXiv:2408.00714}, 2024.

\bibitem[Rodin et~al.(2024)Rodin, Furnari, Min, Tripathi, and Farinella]{rodin2024action}
Ivan Rodin, Antonino Furnari, Kyle Min, Subarna Tripathi, and Giovanni~Maria Farinella.
\newblock Action scene graphs for long-form understanding of egocentric videos.
\newblock In \emph{Proceedings of the IEEE/CVF Conference on Computer Vision and Pattern Recognition}, pages 18622--18632, 2024.

\bibitem[Sarkar et~al.(2025)Sarkar, Miksik, Pollefeys, Barath, and Armeni]{sarkar2025crossover}
Sayan~Deb Sarkar, Ondrej Miksik, Marc Pollefeys, Daniel Barath, and Iro Armeni.
\newblock Crossover: 3d scene cross-modal alignment.
\newblock In \emph{Proceedings of the Computer Vision and Pattern Recognition Conference}, pages 8985--8994, 2025.

\bibitem[Shea et~al.(2017)Shea, Fu, Sun, Cai, Ma, Fan, Gong, and Liu]{shea2017location}
Ryan Shea, Di~Fu, Andy Sun, Chao Cai, Xiaoqiang Ma, Xiaoyi Fan, Wei Gong, and Jiangchuan Liu.
\newblock Location-based augmented reality with pervasive smartphone sensors: Inside and beyond pokemon go!
\newblock \emph{IEEE Access}, 5:\penalty0 9619--9631, 2017.

\bibitem[Shi et~al.(2025)Shi, Jain, Chi, Doh, Chi, Quinn, and Ramani]{shi2025caring}
Jingyu Shi, Rahul Jain, Seunggeun Chi, Hyungjun Doh, Hyung-gun Chi, Alexander~J Quinn, and Karthik Ramani.
\newblock Caring-ai: Towards authoring context-aware augmented reality instruction through generative artificial intelligence.
\newblock In \emph{Proceedings of the 2025 CHI Conference on Human Factors in Computing Systems}, pages 1--23, 2025.

\bibitem[Sumers et~al.(2023)Sumers, Yao, Narasimhan, and Griffiths]{sumers2023cognitive}
Theodore Sumers, Shunyu Yao, Karthik Narasimhan, and Thomas Griffiths.
\newblock Cognitive architectures for language agents.
\newblock \emph{Transactions on Machine Learning Research}, 2023.

\bibitem[Takacs et~al.(2008)Takacs, Chandrasekhar, Gelfand, Xiong, Chen, Bismpigiannis, Grzeszczuk, Pulli, and Girod]{takacs2008outdoors}
Gabriel Takacs, Vijay Chandrasekhar, Natasha Gelfand, Yingen Xiong, Wei-Chao Chen, Thanos Bismpigiannis, Radek Grzeszczuk, Kari Pulli, and Bernd Girod.
\newblock Outdoors augmented reality on mobile phone using loxel-based visual feature organization.
\newblock In \emph{Proceedings of the 1st ACM international conference on Multimedia information retrieval}, pages 427--434, 2008.

\bibitem[Teng et~al.(2021)Teng, Wang, Li, and Wu]{teng2021target}
Yao Teng, Limin Wang, Zhifeng Li, and Gangshan Wu.
\newblock Target adaptive context aggregation for video scene graph generation.
\newblock In \emph{Proceedings of the IEEE/CVF International Conference on Computer Vision}, pages 13688--13697, 2021.

\bibitem[Vir and Madinei(2024)]{vir2024archef}
Rithik Vir and Parsa Madinei.
\newblock Archef: An ios-based augmented reality cooking assistant powered by multimodal gemini llm.
\newblock \emph{arXiv preprint arXiv:2412.00627}, 2024.

\bibitem[Wang et~al.(2024{\natexlab{a}})Wang, Hanson, Li, Papakonstantinou, Simhadri, and Xie]{wang2024vector}
Jianguo Wang, Eric Hanson, Guoliang Li, Yannis Papakonstantinou, Harsha Simhadri, and Charles Xie.
\newblock Vector databases: What's really new and what's next?(vldb 2024 panel).
\newblock \emph{Proceedings of the VLDB Endowment}, 17\penalty0 (12):\penalty0 4505--4506, 2024{\natexlab{a}}.

\bibitem[Wang et~al.(2024{\natexlab{b}})Wang, Yu, Zhao, Sun, Hou, Liang, Hu, Han, and Gan]{wang2024karma}
Zixuan Wang, Bo~Yu, Junzhe Zhao, Wenhao Sun, Sai Hou, Shuai Liang, Xing Hu, Yinhe Han, and Yiming Gan.
\newblock Karma: Augmenting embodied ai agents with long-and-short term memory systems.
\newblock \emph{arXiv preprint arXiv:2409.14908}, 2024{\natexlab{b}}.

\bibitem[Wang et~al.(2024{\natexlab{c}})Wang, Mao, Fried, and Neubig]{wang2024agent}
Zora~Zhiruo Wang, Jiayuan Mao, Daniel Fried, and Graham Neubig.
\newblock Agent workflow memory.
\newblock \emph{arXiv preprint arXiv:2409.07429}, 2024{\natexlab{c}}.

\bibitem[Wu et~al.(2024)Wu, Qian, Castelo~Quispe, Chen, Rulff, and Silva]{wu2024artist}
Guande Wu, Jing Qian, Sonia Castelo~Quispe, Shaoyu Chen, Jo{\~a}o Rulff, and Claudio Silva.
\newblock Artist: Automated text simplification for task guidance in augmented reality.
\newblock In \emph{Proceedings of the 2024 CHI Conference on Human Factors in Computing Systems}, pages 1--24, 2024.

\bibitem[Xu et~al.(2024)Xu, Nguyen, and Du]{xu2024augmented}
Fang Xu, Tri Nguyen, and Jing Du.
\newblock Augmented reality for maintenance tasks with chatgpt for automated text-to-action.
\newblock \emph{Journal of Construction Engineering and Management}, 150\penalty0 (4):\penalty0 04024015, 2024.

\bibitem[Yang et~al.(2025{\natexlab{a}})Yang, Guo, Xu, Yan, Chen, Xing, and Jiang]{yang2025socialmind}
Bufang Yang, Yunqi Guo, Lilin Xu, Zhenyu Yan, Hongkai Chen, Guoliang Xing, and Xiaofan Jiang.
\newblock Socialmind: Llm-based proactive ar social assistive system with human-like perception for in-situ live interactions.
\newblock \emph{Proceedings of the ACM on Interactive, Mobile, Wearable and Ubiquitous Technologies}, 9\penalty0 (1):\penalty0 1--30, 2025{\natexlab{a}}.

\bibitem[Yang et~al.(2025{\natexlab{b}})Yang, Kim, Kim, Kwak, Park, Hong, Woo, and Yeo]{yang2025llm}
Dongil Yang, Minjin Kim, Sunghwan Kim, Beong-woo Kwak, Minjun Park, Jinseok Hong, Woontack Woo, and Jinyoung Yeo.
\newblock Llm meets scene graph: Can large language models understand and generate scene graphs? a benchmark and empirical study.
\newblock \emph{arXiv preprint arXiv:2505.19510}, 2025{\natexlab{b}}.

\bibitem[Zauner et~al.(2003)Zauner, Haller, Brandl, and Hartman]{zauner2003authoring}
J{\"u}rgen Zauner, Michael Haller, Alexander Brandl, and Werner Hartman.
\newblock Authoring of a mixed reality assembly instructor for hierarchical structures.
\newblock In \emph{The Second IEEE and ACM International Symposium on Mixed and Augmented Reality, 2003. Proceedings.}, pages 237--246. IEEE, 2003.

\bibitem[Zhang et~al.(2024)Zhang, Chen, and Liao]{zhang2024sgedit}
Zhiyuan Zhang, DongDong Chen, and Jing Liao.
\newblock Sgedit: Bridging llm with text2image generative model for scene graph-based image editing.
\newblock \emph{arXiv preprint arXiv:2410.11815}, 2024.

\bibitem[Zheng et~al.(2024)Zheng, Chen, He, Gu, Li, Yang, Lin, Wang, Wang, and Wang]{zheng2024editroom}
Kaizhi Zheng, Xiaotong Chen, Xuehai He, Jing Gu, Linjie Li, Zhengyuan Yang, Kevin Lin, Jianfeng Wang, Lijuan Wang, and Xin~Eric Wang.
\newblock Editroom: Llm-parameterized graph diffusion for composable 3d room layout editing.
\newblock \emph{arXiv preprint arXiv:2410.12836}, 2024.

\bibitem[Zhou et~al.(2024)Zhou, Zhu, Caesar, and Shi]{zhou2024openpsg}
Zijian Zhou, Zheng Zhu, Holger Caesar, and Miaojing Shi.
\newblock Openpsg: Open-set panoptic scene graph generation via large multimodal models.
\newblock In \emph{European Conference on Computer Vision}, pages 199--215. Springer, 2024.

\bibitem[Zhu et~al.(2024)Zhu, Ou, Mou, and Tang]{zhu2024retrieval}
Yichen Zhu, Zhicai Ou, Xiaofeng Mou, and Jian Tang.
\newblock Retrieval-augmented embodied agents.
\newblock In \emph{Proceedings of the IEEE/CVF Conference on Computer Vision and Pattern Recognition}, pages 17985--17995, 2024.

\bibitem[Zitkovich et~al.(2023)Zitkovich, Yu, Xu, Xu, Xiao, Xia, Wu, Wohlhart, Welker, Wahid, et~al.]{zitkovich2023rt}
Brianna Zitkovich, Tianhe Yu, Sichun Xu, Peng Xu, Ted Xiao, Fei Xia, Jialin Wu, Paul Wohlhart, Stefan Welker, Ayzaan Wahid, et~al.
\newblock Rt-2: Vision-language-action models transfer web knowledge to robotic control.
\newblock In \emph{Conference on Robot Learning}, pages 2165--2183. PMLR, 2023.

\end{thebibliography}
